\documentclass{article}

\usepackage{microtype}
\usepackage{graphicx}
\usepackage{subcaption}
\usepackage{booktabs} 

\usepackage{hyperref}

\usepackage[accepted]{icml2026}

\usepackage{amsmath}
\usepackage{amssymb}
\usepackage{mathtools}
\usepackage{amsthm}

\usepackage{algorithm}
\usepackage{algorithmic}

\usepackage{multirow}

\usepackage[capitalize,noabbrev]{cleveref}

\theoremstyle{plain}

\theoremstyle{definition}

\theoremstyle{remark}

\usepackage[textsize=tiny]{todonotes}

\icmltitlerunning{PointCHR: Point Cloud Analysis via Curvature-Aware Hyperbolic Rectification}

\begin{document}

\twocolumn[
  \icmltitle{PointCHR: Point Cloud Analysis via Curvature-Aware Hyperbolic Rectification}



  \icmlsetsymbol{equal}{*}
  \begin{icmlauthorlist}
    \icmlauthor{Xinxing Yu}{MUST}
    \icmlauthor{Liying Yang}{MUST}
    \icmlauthor{Hao Mo}{MUST}
    \icmlauthor{Hui Ma}{equal,GBU,TKU}
    \icmlauthor{Fang Kai}{MUST}
    \icmlauthor{Ajian Liu}{MUST,MAIS}
    \icmlauthor{Yanyan Liang}{equal,MUST}
  \end{icmlauthorlist}

  \icmlaffiliation{MUST}{School of Computer Science and Engineering, Macau University of Science and Technology, Macau, China}
  \icmlaffiliation{GBU}{School of Computing and Information Technology, Great Bay University, Dongguan, China}
  \icmlaffiliation{MAIS}{MAIS, The Institute of Automation of the Chinese Academy of Sciences, Beijing, China}
  \icmlaffiliation{TKU}{Shenzhen International Graduate School, Tsinghua University, Shenzhen, China}

  \icmlcorrespondingauthor{Hui Ma}{mahuilight@gmail.com}
  \icmlcorrespondingauthor{Yanyan Liang}{yyliang@must.edu.mo}

  \icmlkeywords{Machine Learning, ICML}

  \vskip 0.3in
]



\printAffiliationsAndNotice{\icmlEqualContribution}

\begin{abstract}
  High-curvature regions in 3D point clouds encapsulate critical fine-grained geometric semantics yet exhibit a distinct long-tail sparsity in their spatial distribution. 
  The inherent limitations of polynomial volume growth in Euclidean space frequently render these intricate geometric features challenging to adequately resolve within a uniform-scale feature space. 
  Consequently, these regions are frequently overshadowed by smooth global features dominated by low-curvature regions, thereby limiting the discriminative capacity of the network. 
  To address this issue, we propose PointCHR, a curvature-aware hyperbolic rectification (CHR) for point cloud analysis. 
  Utilising the property of exponential volume expansion in the vicinity of hyperbolic manifolds, CHR presents a learnable curvature-guided radial rectification mechanism. 
  By adaptively projecting high-curvature points towards boundary regions endowed with larger effective embedding capacities, PointCHR effectively mitigates the representation crowding problem inherent in Euclidean settings. 
  Extensive experimentation has demonstrated that PointCHR significantly enhances the ability of backbone to capture fine-grained geometric details, achieving state-of-the-art performance across multiple benchmarks.
\end{abstract}

\section{Introduction}

\begin{figure}[ht]
  \vskip 0.2in
  \begin{center}
    \centerline{\includegraphics[width=\linewidth]{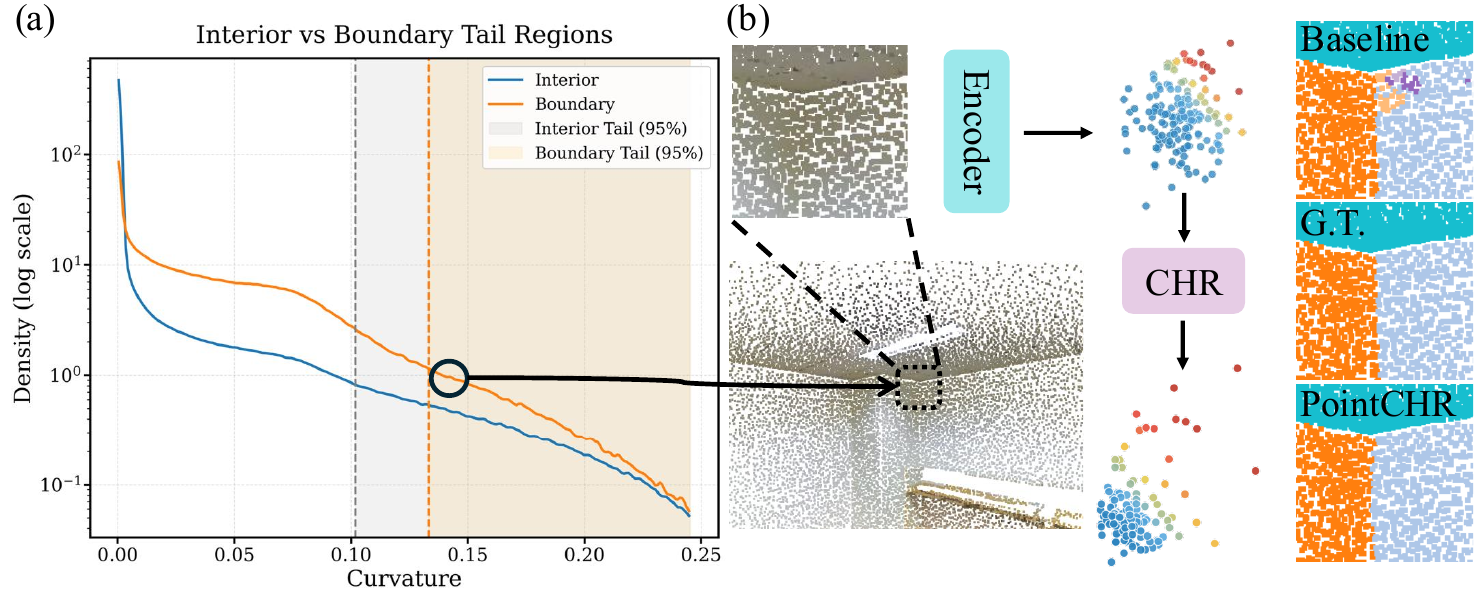}}
    \caption{
        Illustration of the curvature imbalance challenge. (a) Quantitative analysis shows the long-tailed nature of curvature distribution in S3DIS datast. (b) In complex corner regions, conventional encoders fail to handle such tail samples, leading to boundary artifacts and noise. Our CHR addresses this by incorporating geometric priors for feature optimization, successfully restoring crisp boundaries consistent with the Ground Truth.
        }
    \label{fig:longtail_analysis}
  \end{center}
\end{figure}


3D understanding serves as the cornerstone for numerous applications, such as detection\cite{Liu_2025_ICCV,schreier2023offline},reconstruction \cite{Zhu_2023_ICCV,zhu2023garnet,Yang_2023_ICCV,yu2025facnet,wang2026mcgs} and generation \cite{Yang_2025_ICCV,yang2026spatial,han2024pmmtalk,han2025pestalk}. Recent advances in point cloud deep learning have led to a series of effective architectures \cite{qian_pointnext_2022,wang_pointramba_2024}, which show capability in learning global representations. Notwithstanding these advances, most methods aggregate local neighborhoods within Euclidean feature spaces and implicitly presume a homogeneous and universally adequate representational capacity across all geometric primitives.

However, the prevailing Euclidean assumption overlooks the intrinsic heterogeneity of three dimension (3D) geometries. As shown in \cref{fig:longtail_analysis}(a), local curvature exhibits a long-tail distribution: most points lie on low-curvature surfaces , whereas high-curvature structures, such as corners, edges, and fine-grained details are sparsely distributed. Although rare, these regions often carry informative fine-grained geometric cues vital for accurate recognition.

This geometric heterogeneity imposes a structural impediment on Euclidean embeddings employing isotropic scaling. Constrained by the polynomial volume growth of Euclidean space, sparse yet topologically complex high-curvature structures are compelled to vie for a finite representational budget against ubiquitous low-curvature surfaces. This capacity imbalance precipitates representation crowding, wherein fine-grained geometric cues are compressed into restricted embedding subspaces and subsumed by dominant global patterns. As evidenced in \cref{fig:longtail_analysis}(b), such crowding induces a feature collapse of high-curvature regions into indistinguishable latent neighborhoods, thereby eroding class separability and compromising boundary localization. These observations indicate that the fundamental limitation of existing methods stems from the inefficient allocation of representational resources by isotropic embedding strategies across regions of varying geometric density. This necessitates a geometry-aware reparameterization mechanism to adaptively redistribute representational capacity commensurate with local geometric complexity, thereby fundamentally circumventing the crowding bottleneck.

To surmount this bottleneck, we propose aligning the feature space with the intrinsic geometric complexity of the data. Hyperbolic geometry, characterized by constant negative curvature, provides an optimal inductive bias for non-uniform structures. Diverging from Euclidean space and its polynomial growth limitations, hyperbolic space facilitates exponential volume expansion. This property generates immense representational capacity near the boundary and enables the model to unfold crowded high-curvature details into distinct high-resolution regions while preserving global structural integrity. By projecting complex geometric cues toward the boundary while retaining simpler features near the center, hyperbolic embeddings naturally accommodate the long-tailed geometric complexity of 3D point clouds and provide a principled solution to the crowding problem.

However, this direct adaptation is hindered by three structural impediments. First, structural heterogeneity: as shown in \cref{fig:longtail_analysis}(a), the semantic ambiguity arising from the entanglement of object boundaries and fine textures within high-curvature regions renders static curvature priors ineffective. Second, feature misalignment: the discrepancy between Euclidean features and the intrinsic geometric metric of the manifold restricts the effective activation of hyperbolic hierarchical capacity. Third, optimization instability: the asymptotic nature of boundary regions predisposes tangent-space transformations to gradient vanishing or numerical overflow, obstructing stable model convergence.

To address these challenges, we propose \textbf{PointCHR}, a curvature aware hyperbolic rectification for point cloud analysis comprising three integral components, namely Hyperbolic Semantic Transformation (\textbf{HST}), Point-wise Curvature-Adaptive Perception (\textbf{PCP}), and Closed-Form Geodesic Dilation (\textbf{CGD}). 
Specifically, targeting structural heterogeneity, PCP discards static heuristics in favor of leveraging curvature as geometric guidance fused with local context to dynamically learn point-wise radial scaling factors, which achieves a data-driven and discriminative disentanglement of object boundaries and fine textures. 
Subsequently, to resolve feature misalignment, HST projects Euclidean backbone features onto the Poincaré ball and employs Möbius linear transformations to reconstruct the semantic space on the manifold, thereby explicitly aligning the feature distribution with intrinsic hyperbolic geometry. 
Finally, to surmount optimization instability, CGD directly modulates geodesic distances based on the learned scaling factors via a tangent-space-free closed-form solution. This mechanism effectively circumvents the risk of numerical collapse in asymptotic boundary regions and enables the model to robustly unfold crowded high-curvature features into the high-capacity regions of the hyperbolic space.

In summary, our contributions are as follows:
\begin{itemize}
  \item To the best of our knowledge, PointCHR constitutes the pioneering attempt to synergize intrinsic curvature cues with hyperbolic feature learning for point cloud analysis. This effectively bridges the gap between geometric topology and representation learning to offer a principled solution for representation crowding.

  \item We design a unified rectification pipeline integrating HST, PCP, and CGD to systematically resolve structural impediments. These components collaboratively correct feature misalignment, disentangle structural heterogeneity via adaptive scaling, and ensure optimization stability near the asymptotic boundary.

  \item Extensive experiments demonstrate that our method achieves state-of-the-art (SOTA) performance on standard benchmarks. Furthermore, qualitative results verify that PointCHR significantly enhances boundary delineation in geometrically complex regions.
\end{itemize}

\section{Related Work}

\noindent \textbf{3D Point Cloud Analysis.} Modern approaches generally follow a sample-and-group paradigm to aggregate local contexts. Mainstream backbones encode neighborhoods via convolution-based operators \cite{zhang_point-m2ae_2022,zhang_pcp-mae_2024,zhang_towards_2025,lin_meta_2023,deng_pointvector_2023,deng_linnet_2024,zheng_point_2024,li_pointgl_2024,zha_point_2025,zha2025exploring,zeng_deepla-net_2025}, graph attention \cite{wang_dynamic_2019,xu_paconv_2021,xu2021rpvnet,yin_point_2024,koch_open3dsg_2024}, Transformer \cite{vaswani_attention_2017} architectures \cite{zhao_point_2021,park_fast_2022,wu_point_2022,park_self-positioning_2023,luo2024exploring,wu_point_2024,wu_sonata_2025,thomas_kpconvx_2024,xu_group_2024,kolodiazhnyi_oneformer3d_2024,yang_swin3d_2025,han_all_2025}, or recent Mamba-based models \cite{gu_mamba_2024,liang_pointmamba_2024,han_mamba3d_2024,zhang_point_2025,li_feast_2025,bahri_spectral_2025,lin_2025_zigzagpointmamba,yu_2026_aaai,yu2026pointcsp}. To further enhance geometric sensitivity, several works incorporate explicit priors. RepSurf \cite{Ran_2022_CVPR}, CurveNet \cite{Xiang_2021_ICCV} and CanonNet \cite{friedmann2025canonnet} augment inputs with attributes like curvature, while RS-CNN \cite{liu2019relation} and Point-NN \cite{zhang2023point} exploit topological relations. Others employ geometry-aware attention \cite{Qin_2022_CVPR, xu2021learning} or boundary-specific contrastive learning \cite{Tang_2022_CVPR} to focus on complex regions.
While these methods have achieved success in capturing shape features, they rely on isotropic Euclidean embeddings. However, constrained by the polynomial volume growth of Euclidean space, such approaches struggle to accommodate the expansive information of complex local geometries. This bottleneck induces a representation crowding effect, where fine-grained semantics in high-curvature regions are inevitably submerged by the dominant features of smooth surfaces, limiting the discriminative power of the network on critical boundary details.

\noindent  \textbf{Hyperbolic Deep Learning.} 
Hyperbolic geometry has emerged as a powerful paradigm for modeling data with latent hierarchical structures and exponential volume growth.
Seminal works \cite{nickel2017poincare, ganea2018hyperbolic} introduced Poincaré embeddings and hyperbolic neural networks, demonstrating superior capacity over Euclidean counterparts in preserving semantic hierarchies with lower dimensionality. These advantages have been successfully extended to graph representation learning \cite{chami2019hyperbolic} and computer vision tasks, such as classification \cite{khrulkov2020hyperbolic,montanaro2022rethinking,feng2025hyperbolic} and zero-shot recognition \cite{liu2020hyperbolic}. Recent advances have further integrated hyperbolic geometry into modern architectures. For instance, Hyperbolic Vision Transformers \cite{ermolov2022hyperbolic} combine Euclidean and hyperbolic features to capture diverse visual semantics, while optimization strategies like clipped hyperbolic classifiers \cite{Guo_2022_CVPR} have been proposed to enhance training stability. However, the application of hyperbolic geometry to 3D point cloud analysis remains underexplored. Specifically, leveraging the exponential capacity of hyperbolic boundary regions to alleviate the feature crowding of high-curvature geometric primitives remains a promising yet underexplored avenue in 3D analysis.

\section{Preliminaries}
\label{sec:preliminaries}

\subsection{Curvature in Point Clouds}

Given a point cloud $\mathcal{P}=\{\mathbf{p}_i\in\mathbb{R}^3\}_{i=1}^{N}$, the local geometric complexity around each point can be characterized by the variation of its spatial neighborhood. We adopt the widely used Principal Component Analysis \cite{hotelling1933analysis} approach to estimate point-wise curvature \cite{pauly2002surface}.

For each $\mathbf{p}_i$, we form the centered neighborhood matrix $\mathbf{X}_i \in \mathbb{R}^{k \times 3}$ using its $k$-nearest neighbors $\mathcal{N}(i)$:
\begin{equation}
    \mathbf{X}_i = \left[ (\mathbf{p}_{j_1} - \mathbf{p}_i), \dots, (\mathbf{p}_{j_k} - \mathbf{p}_i) \right]^\top.
\end{equation}
The local geometric structure is encoded in the covariance matrix $\mathbf{C}_i = \frac{1}{k}\mathbf{X}_i^\top \mathbf{X}_i \in \mathbb{R}^{3\times 3}$. Let $\lambda_{i,1} \ge \lambda_{i,2} \ge \lambda_{i,3} \ge 0$ denote the eigenvalues of $\mathbf{C}_i$. The smallest eigenvalue, $\lambda_{i,3}$, quantifies the variance along the surface normal, thereby reflecting the deviation of the local neighborhood from a planar surface.
Following established conventions, we calculate the curvature estimate $\kappa_i$ at point $\mathbf{p}_i$ as:
\begin{equation}
    \kappa_i = \frac{\lambda_{i,3}}{\sum_{m=1}^3 \lambda_{i,m} + \varepsilon},
    \label{eq:pca_curvature}
\end{equation}
where $\varepsilon$ is a small constant for numerical stability.

\subsection{Hyperbolic Geometry Basics}

We employ the Poincaré ball model to represent hyperbolic geometry, as it is well-suited for gradient-based optimization.
The Poincaré ball $(\mathbb{B}_c^d, g^\mathbb{B})$ of dimension $d$ with manifold curvature $-c$ ($c > 0$) is defined as the manifold $\mathbb{B}_c^d = \{ \mathbf{x} \in \mathbb{R}^d : \| \mathbf{x} \|^2 < 1/c \}$, equipped with the Riemannian metric $g_\mathbf{x}^\mathbb{B} = \lambda_\mathbf{x}^2 g^E$, where $\lambda_\mathbf{x} = \frac{2}{1 - c\|\mathbf{x}\|^2}$ is the conformal factor and $g^E$ denotes the Euclidean metric.

While the theoretical domain is the open ball $\mathbb{B}_c^d$, numerical implementations require bounding the inputs away from the boundary to prevent arithmetic overflow, since $\text{arctanh}(z) \to \infty$ as $z \to 1$. 
We therefore operate within a closed subset $\mathbb{B}_{c, \varepsilon}^d \subset \mathbb{B}_{c}^d$:
\begin{equation}
    \mathbb{B}_{c, \varepsilon}^d = \left\{ \mathbf{x} \in \mathbb{R}^d \mid \|\mathbf{x}\| \le \frac{1 - \varepsilon}{\sqrt{c}} \right\},
\end{equation}
where $\varepsilon$ is a small constant margin.


An input feature vector $\mathbf{v} \in \mathbb{R}^d$ is mapped to the Poincaré ball via the exponential map $\text{Exp}_{\mathbf{0}}^c$, and mapped back via the logarithmic map $\text{Log}_{\mathbf{0},\varepsilon}^c$.
The numerically stable formulations are given by:
\begin{align}
    \text{Exp}_{\mathbf{0}}^c(\mathbf{v}) &= \tanh(\sqrt{c}\|\mathbf{v}\|) \frac{\mathbf{v}}{\sqrt{c}\|\mathbf{v}\|}, \\
    \text{Log}_{\mathbf{0},\varepsilon}^c(\mathbf{x}) &= \text{arctanh}(\sqrt{c}\|\tilde{\mathbf{x}}\|) \frac{\tilde{\mathbf{x}}}{\sqrt{c}\|\tilde{\mathbf{x}}\|}
\end{align}
where $\tilde{\mathbf{x}} = \min\left(1, \frac{1 - \varepsilon}{\sqrt{c} \|\mathbf{x}\|}\right) \mathbf{x}.$
Note that for $\mathbf{v} \to \mathbf{0}$, the term $\frac{\tanh(v)}{v} \to 1$, ensuring the exponential map approximates the identity mapping near the origin.
This mechanism allows us to lift standard Euclidean features onto the hyperbolic manifold to leverage its geometric properties.

\section{Method}
\label{sec:method}

\subsection{Overall Architecture}
\label{subsec:overall}

\begin{figure*}[ht]
  \vskip 0.2in
  \begin{center}
    \centerline{\includegraphics[width=\linewidth]{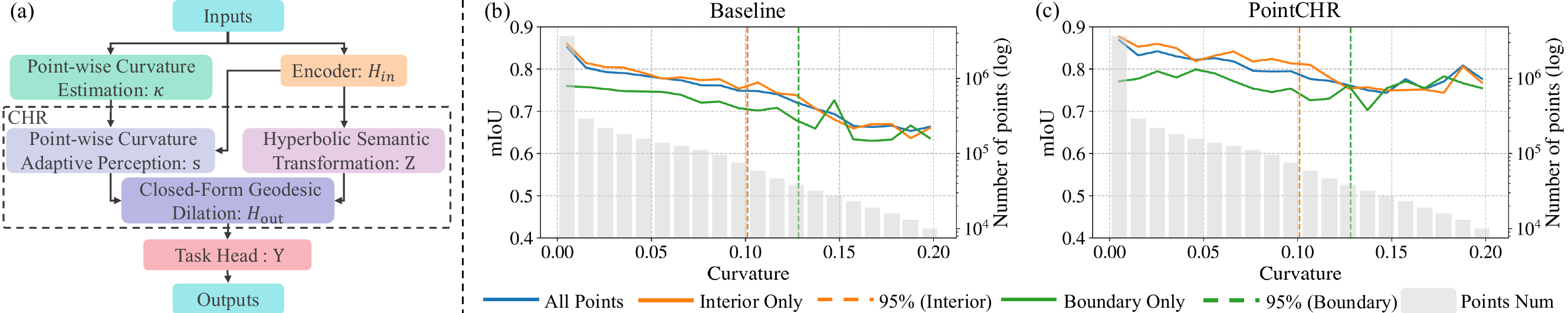}}
    \caption{
        \textbf{(a)} Overview of the proposed PointCHR framework.
        \textbf{(b)} The Euclidean baseline struggles with this data imbalance, showing a sharp performance degradation in the high-curvature tail. 
        \textbf{(c)} Our PointCHR effectively rectifies this issue. By leveraging hyperbolic geometry to handle structural complexity, our method prevents performance collapse in the tail, demonstrating superior robustness on rare hard samples. The background histograms visualize the long-tail distribution of point curvature in the S3DIS dataset, highlighting the sparsity of high-curvature samples.
        }
    \label{fig:mainfold_tail_analysis}
  \end{center}
\end{figure*}

We formulate our framework as a mapping $\Phi: \{\mathcal{P}, \mathcal{F}\} \to \mathbf{Y}$, transforming the input point cloud $\mathcal{P}$ and auxiliary features $\mathcal{F} \in\mathbb{R}^{N \times M}$ into task-specific predictions $\mathbf{Y}$. As shown in \cref{fig:mainfold_tail_analysis}(a), the pipeline consists of three stages:

\noindent  \textbf{Feature Initialization.} 
We first extract the structural prior $\boldsymbol{\kappa} \in \mathbb{R}^N$ and the initial semantic embedding $\mathbf{H}_{\text{in}} \in \mathbb{R}^{N \times C}$ via a foundational feature encoder $\mathcal{E}$:
\begin{equation}
    \boldsymbol{\kappa} = \Psi_{\text{curv}}(\mathcal{P}), \quad \mathbf{H}_{\text{in}} = \mathcal{E}([\mathcal{P} \oplus \mathcal{F}]),
\end{equation}
where $\Psi_{\text{curv}}$ denotes the point-wise curvature estimation and $\oplus$ indicates channel-wise concatenation.

\noindent  \textbf{Curvature-Aware Hyperbolic Rectification (CHR).}
The deep feature extraction is governed by the \textbf{CHR} $\mathcal{B}$. Formally, the backbone output $\mathbf{H}_{\text{out}}$ is derived as:

\begin{equation}
    \mathbf{H}_{\text{out}} = \mathcal{B}(\mathbf{H}_{\text{in}} \mid \boldsymbol{\kappa}) \in \mathbb{R}^{N \times C'}.
\end{equation}
The notation $(\cdot \mid \boldsymbol{\kappa})$ emphasizes that the deep feature transformation is strictly conditioned on the geometric prior $\boldsymbol{\kappa}$.

\noindent  \textbf{Task Prediction.}
Finally, the geometrically rectified representations $\mathbf{H}_{\text{out}}$ are projected onto the target label space via a task-specific decoder $\mathcal{T}$. Whether for classification or segmentation, the prediction $\mathbf{Y}$ is generated as:
\begin{equation}
    \mathbf{Y} = \mathcal{T}(\mathbf{H}_{\text{out}}).
\end{equation}


\subsection{Curvature-Aware Hyperbolic Rectification (CHR)}
\label{subsec:CHR}

Standard Euclidean networks often struggle to distinguish features in regions with high geometric complexity due to the lack of sufficient margin. To address this, we propose the CHR, which dynamically adjusts the hyperbolic embeddings based on local point geometry cues. The block consists of three stages: Hyperbolic Semantic Transformation, Point-wise Curvature-Adaptive Perception, and Closed-Form Geodesic Rescaling. The overall procedure is summarized in \cref{alg:ghr_block}.

\noindent  \textbf{Hyperbolic Semantic Transformation (HST).}
Let $\mathbf{h}^{in}_i \in \mathbb{R}^{C}$ denote the input feature vector for the $i$-th point in the ambient Euclidean space. 
To leverage the hyperbolic geometry, we first lift the input onto the Poincaré ball. 
By identifying the Euclidean input space with the tangent space at the origin ($T_{\mathbf{0}}\mathbb{B}_c^{C} \cong \mathbb{R}^{C}$), we apply the exponential map:
\begin{equation}
    \mathbf{z}_i^{(0)} = \text{Exp}_{\mathbf{0}}^c(\mathbf{h}^{in}_i) \in \mathbb{B}_c^{C}.
\end{equation}
This operation projects the standard features onto the manifold, initializing the hyperbolic embedding process.

To facilitate expressive semantic feature extraction while strictly preserving the underlying hyperbolic geometry, we employ the M\"obius linear layer.
Standard Euclidean linear transformations are fundamentally incompatible with this objective, as they disregard the underlying conformal geometry of the Poincaré ball, thereby inevitably introducing severe geometric distortions.
Consequently, we adopt the M\"obius generalization of the linear map, which is defined via M\"obius matrix multiplication ($\otimes_c$) and addition ($\oplus_c$):
\begin{equation}
    \mathbf{z}'_i = \text{M\"obiusLinear}(\mathbf{z}_i^{(0)}) = (W \otimes_c \mathbf{z}_i^{(0)}) \oplus_c \mathbf{b},
\end{equation}
where $W \in \mathbb{R}^{C \times C}$ is the learnable weight matrix and $\mathbf{b} \in \mathbb{B}_c^{C}$ is the hyperbolic bias.

To introduce non-linearity, we apply the M\"obius variant of the GELU activation \cite{hendrycks2016gaussian}, denoted as $\sigma^{\otimes_c}$. 
This operation ensure the resulting features remain well-defined on the manifold:
\begin{equation}
    \mathbf{z}_i^{(1)} = \text{M\"obiusActivation}(\mathbf{z}_i^{'}) =\sigma^{\otimes_c}(\mathbf{z}'_i).
\end{equation}
By stacking these operations, the block effectively learns complex semantic interactions without degrading the hierarchical structure encoded in the hyperbolic embeddings.

\begin{algorithm}[tb]
    \caption{Curvature-Aware Hyperbolic Rectification}
    \label{alg:ghr_block}
    \begin{algorithmic}[1]
        \STATE {\bfseries Input:} Euclidean feature matrix $\mathbf{H}_{\text{in}} \in \mathbb{R}^{N \times C}$, Normalized point curvature $\boldsymbol{\kappa} \in \mathbb{R}^{N}$
        \STATE {\bfseries Hyperparameters:} Manifold curvature $c$, Dilation scale $\alpha$, Sensitivity $\gamma$, Stability margin $\varepsilon$
        \STATE {\bfseries Output:} Rectified Euclidean features $\mathbf{H}_{\text{out}} \in \mathbb{R}^{N \times C}$
        
        \STATE \textit{// 1. Hyperbolic Semantic Transformation}
        \STATE $\mathbf{Z}^{(0)} \leftarrow \text{Exp}_{\mathbf{0}}^c(\mathbf{H}_{\text{in}})$
        \STATE $\mathbf{Z}^{(1)} \leftarrow \text{M\"obiusActivation}(\text{M\"obiusLinear}(\mathbf{Z}^{(0)}))$
        
        \STATE \textit{// 2. Point-wise Curvature-Adaptive Perception}
        \STATE $\mathbf{G} \leftarrow \text{Softplus}(\text{MLP}(\mathbf{H}_{\text{in}} \oplus \boldsymbol{\kappa}))$ 
        \STATE $\mathbf{s} \leftarrow 1 + \alpha \cdot \mathbf{G} \odot (\boldsymbol{\kappa})^\gamma$ 
        
        \STATE \textit{// 3. Closed-Form Geodesic Dilation}
        \STATE Initialize $\mathbf{Z}' \in \mathbb{B}_c^{N \times C}$
        \FOR{each point $i \in \{1, \dots, N\}$}
            \STATE $\mathbf{z}_i \leftarrow \mathbf{Z}^{(1)}_i$ 
            \STATE $\tilde{r}_i \leftarrow \max(\|\mathbf{z}_i\|_2, \varepsilon)$ 
            \STATE $d'_i \leftarrow s_i \cdot \text{arctanh}(\sqrt{c} \cdot \tilde{r}_i)$ 
            \STATE $\lambda_i \leftarrow \frac{1}{\sqrt{c}\tilde{r}_i} \tanh(d'_i)$
            \STATE $\mathbf{Z}'_i \leftarrow \lambda_i \cdot \mathbf{z}_i$ 
        \ENDFOR
        
        \STATE \textit{// 4. Projection back to Euclidean Space}
        \STATE $\mathbf{H}_{\text{out}} \leftarrow \text{LayerNorm}(\text{Log}_{\mathbf{0},\varepsilon}^c(\mathbf{Z}'))$
        \STATE \textbf{return} $\mathbf{H}_{\text{out}}$
    \end{algorithmic}
\end{algorithm}

\noindent \textbf{Point-wise Curvature-Adaptive Perception (PCP).}
While the preceding semantic transformation effectively captures feature interactions, it operates in an isotropic manner, treating the hyperbolic space as a homogeneous medium. This overlooks a fundamental property of the Poincaré ball: exponential volume expansion. 
Specifically, the spatial capacity of the manifold increases exponentially as the distance from the origin grows. Consequently, regions near the boundary offer significantly larger embedding volume, making them ideal for resolving fine-grained details, whereas the origin is constrained and better suited for low-entropy information.

We posit that the local geometry of 3D point clouds exhibits a latent hierarchy analogous to this spatial structure. 
Points situated on flat surfaces typically represent generic shape structures with low information density. Conversely, points in high-curvature regions encapsulate specific, high-frequency geometric details. 
To maximize representation efficiency, we propose to explicitly align this geometric complexity with the radial capacity of the manifold. Our objective is to distinguishably embed these features by pushing high-curvature points towards the spacious boundary while retaining flat points near the stable origin.
Therefore, we introduce a mechanism to dynamically modulate the embedding of radial position based on point curvature cues. 

Let $\kappa_i \in [0, 1]$ be the normalized local geometric curvature of the $i$-th point. Relying solely on raw point-wise curvature is insufficient, as the importance of a geometric feature often depends on its semantic context. 
Therefore, we propose a learnable gating function $\mathcal{G}_\phi$ to adaptively fuse the geometric cue with the Euclidean semantic feature $\mathbf{h}^{in}_i$. 
The point-wise curvature-aware gating factor $g_i$ is computed as:
\begin{equation}
    g_i = \text{Softplus} \left( \mathcal{G}_\phi([\mathbf{h}^{in}_i \oplus \kappa_i]) \right),
\end{equation}
where $\oplus$ denotes channel-wise concatenation, and $\mathcal{G}_\phi$ is a lightweight learnable Multi-Layer Perceptron.

We then formulate the hierarchy scaling factor $s_i$, which determines the degree of radial dilation. 
This factor couples the learned semantic weight $g_i$ with the geometric prior:
\begin{equation}
    \label{eq:scaling_factor}
    s_i = 1 + \alpha \cdot g_i \cdot (\kappa_i)^\gamma.
\end{equation}
Here, $\alpha > 0$ is a hyperparameter controlling the maximum dilation amplitude, and $\gamma \ge 1$ modulates the sensitivity to point-wise curvature variations. 

\noindent \textbf{Closed-Form Geodesic Dilation (CGD).}
Upon computing the hierarchy scaling factor $s_i$, we adjust the radial position of the embedding to reflect the associated geometric hierarchy. 
We perform this operation via M\"obius scalar multiplication, which scales the geodesic distance from the origin while preserving the angular orientation.

Recall that the geodesic distance between the origin and a point $\mathbf{z} \in \mathbb{B}_c^d$ is given by $d_{\mathbb{B}}(\mathbf{0}, \mathbf{z}) = \frac{1}{\sqrt{c}}\text{arctanh}(\sqrt{c}\|\mathbf{z}\|)$. 
We seek a dilated embedding $\mathbf{z}_i^{(2)}$ that maintains the direction of $\mathbf{z}_i^{(1)}$ but scales the geodesic distance of the vector by the factor $s_i$:
\begin{equation}
    d_{\mathbb{B}}(\mathbf{0}, \mathbf{z}_i^{(2)}) = s_i \cdot d_{\mathbb{B}}(\mathbf{0}, \mathbf{z}_i^{(1)}).
\end{equation}
To implement this efficiently, we leverage the closed-form analytical solution of the M\"obius scalar multiplication:
\begin{equation}
    \label{eq:closed_form}
    \begin{split}
        \mathbf{z}_i^{(2)} &= \frac{1}{\sqrt{c}} \tanh \left( s_i \cdot \text{arctanh}(\sqrt{c} \tilde{r}_i) \right) \frac{\mathbf{z}_i^{(1)}}{\tilde{r}_i},
    \end{split}
\end{equation}
where $\tilde{r}_i = \max(\|\mathbf{z}_i^{(1)}\|, \varepsilon)$ ensures numerical stability.

This formulation provides a direct geometric interpretation of hierarchy adjustment. 
Furthermore, it naturally satisfies the constraints of the manifold: since the image of the $\tanh(\cdot)$ function is strictly $(-1, 1)$, the norm of the rectified embedding $\|\mathbf{z}_i^{(2)}\|$ is intrinsically bounded by $1/\sqrt{c}$. This ensures that the embeddings remain valid within the Poincaré ball throughout the dilation process.

Finally, the rectified embeddings are projected back to Euclidean space for subsequent processing:
\begin{equation}
    \label{eq:log_map}
    \begin{split}
        \mathbf{h}^{out}_i = \text{LayerNorm}(\text{Log}_{\mathbf{0},\varepsilon}^c(\mathbf{z}_i^{(2)})).
    \end{split}
\end{equation}

\section{Experiments}

\subsection{Implementation Details} 
In the proposed CHR, the manifold curvature parameter $c$ is initialized to $1.0$, while the hierarchy scaling hyperparameters are set to $\alpha=1.0$ and $\gamma=2.0$. The numerical stability margin is fixed at $\varepsilon=10^{-5}$. The boundary edge regions radii set as R=0.02.
The model is optimized using the AdamW optimizer~\cite{loshchilov_decoupled_2017} coupled with a cosine annealing learning rate scheduler~\cite{loshchilov2016sgdr}. 
To stabilize the early training dynamics, we incorporate an initial linear warmup phase.
The CamPoint~\cite{zhang_campoint_2025} backbone architecture is adopted as the foundational feature encoder, with the CHR seamlessly integrated after major feature extraction stage.
Crucially, to ensure a strictly fair comparison with baseline methods, we report performance metrics based on standard single-view evaluation, foregoing any multi-view voting mechanisms during inference.
All computational experiments were conducted on a workstation equipped with a single NVIDIA GeForce RTX 4090 GPU.

\subsection{Main Results}

\noindent \textbf{Semantic Segmentation on S3DIS Dataset.}
The S3DIS \cite{armeni_3d_2016} dataset is a widely adopted benchmark for indoor scene understanding. It comprises six large-scale areas with annotated 3D point clouds. Standard evaluation protocols are followed, with results on Area 5 and 6-fold cross-validation reported. The mean Intersection over Union (mIoU) is utilised to quantify segmentation performance.

As illustrated in \cref{tab:S3DIS}, PointCHR achieves a new state-of-the-art performance, attaining an mIoU of 86.0\% on Area 5 and 89.1\% on the 6-fold cross-validation.
This represents significant improvements over prior leading methods, including CamPoint and Sonata, demonstrating the effectiveness of our geometry-aware hyperbolic rectification approach in enhancing semantic segmentation capabilities.



We utilize edge-strip segmentation as a representative stress test. 
Boundary zones serve as the ideal proxy because they represent the intersection of peak geometric complexity and semantic ambiguity.
As shown in \cref{tab:S3DIS_boundaries}, PointCHR achieves consistent improvements in these edge regions.
Furthermore, the holistic SOTA performance reported in \cref{tab:S3DIS} confirms that our method generalizes well beyond just boundaries, effectively rectifying the high-curvature embeddings present in the interior as well.

\begin{figure*}[ht]
  \vskip 0.2in
  \begin{center}
    \centerline{\includegraphics[width=\linewidth]{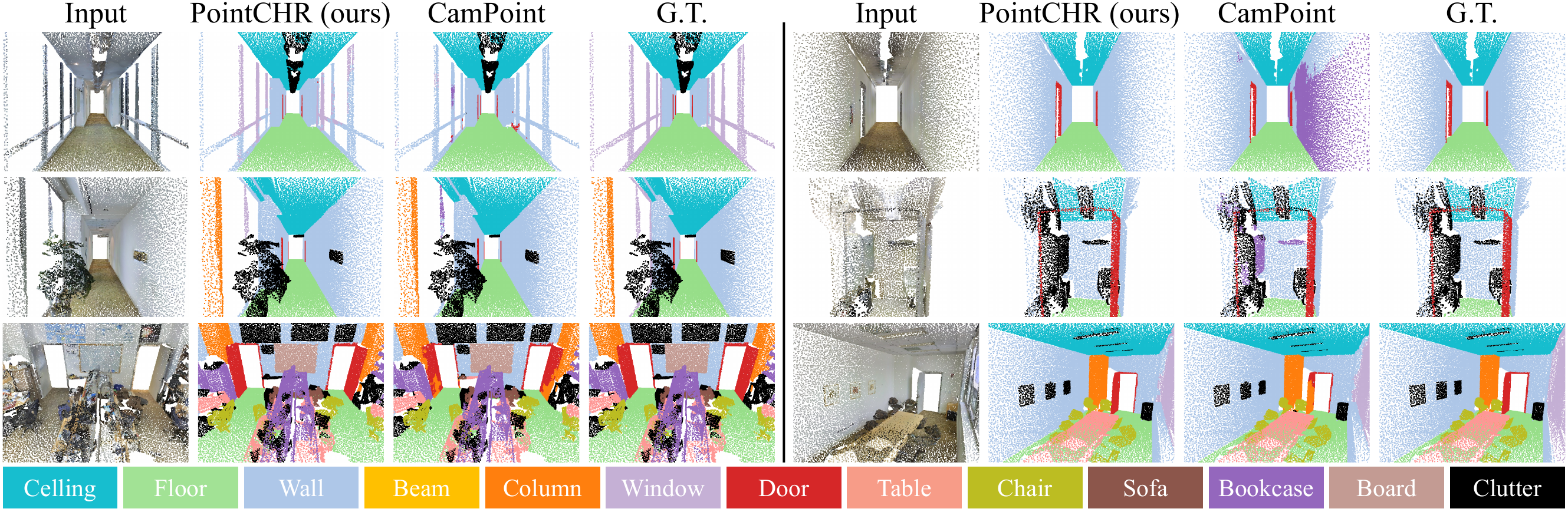}}
    \caption{
        Qualitative comparison of segmentation results on the S3DIS Area5 dataset.
        PointCHR effectively preserves fine-grained boundary details and semantic consistency, significantly outperforming the baseline in challenging scenarios.
        }
    \label{fig:main_results_vis}
  \end{center}
\end{figure*}

\cref{fig:mainfold_tail_analysis} (b) and (c) dissects the impact of geometric complexity on segmentation performance. The background histograms reveal a severe long-tail distribution of curvature, where geometrically complex points are extremely sparse. As shown in \cref{fig:mainfold_tail_analysis}(b), the Euclidean baseline suffers from a performance collapse in this high-curvature tail due to data scarcity. In contrast, \cref{fig:mainfold_tail_analysis}(c) demonstrates the efficacy of our PointCHR. Upon introducing the geometry-aware hyperbolic rectification, the performance curve in the tail region is significantly flattened and uplifted. Notably, PointCHR maintains robust mIoU even in the extreme tail, where the baseline fails. This improvement is consistent across boundary regions and complex interior structures. We attribute this robustness to the exponential expansion capacity of hyperbolic space, which allows our model to embed sparse, high-curvature features with sufficient margin, compensating for the data scarcity in the long-tail distribution. We have observed that our model achieves more significant performance improvements for high-curvature components than for medium-curvature components. We have analyzed this phenomenon in the \textbf{supplementary materials}. The visualization in \cref{fig:main_results_vis} illustrates the qualitative improvements achieved by PointCHR.

\begin{table}[t]
    \caption{Semantic segmentation mIoU results (\%) on the S3DIS dataset are evaluated on Area5 and 6-fold cross-validation.}
    \label{tab:S3DIS}
      \begin{center}
        \begin{small}
          \begin{sc}
            \begin{tabular}{l c c c}
                \toprule
                Method                  &   Ref.        & Area5             & 6-fold      \\
                \midrule
                PTV3                    &   CVPR'24     & 73.4              & 77.7        \\
                HPENet                  &   AAAI'24     & 72.7              & 78.7        \\
                PDNet-XXL               &   AAAI'24     & 72.3              & 78.3        \\ 
                PCM                     &   AAAI'25     & 74.1              & /           \\
                DeepLA-120              &	  CVPR'25     & 75.7              & 79.8        \\
                CamPoint                &   CVPR'25     & \underline{83.3}  & /           \\
                Sonata                  &   CVPR'25     & 76.0              & \underline{82.3}        \\
                VDG                     &   ICCV'25     & 71.5              & 73.2        \\
                Point-PQAE              &   ICCV'25     & 61.4              & /           \\
                \midrule
                PointCHR                &   /           &\textbf{86.0}      &\textbf{89.1}\\
                \bottomrule
            \end{tabular}
      \end{sc}
    \end{small}
  \end{center}
  \vskip -0.1in
\end{table}%

\begin{table}[t]
    \caption{Quantitative comparison of edge-strip segmentation with baseline on S3DIS, reporting mIoU (\%) and F1-scores (\%) under different neighborhood radii (R).}
    \label{tab:S3DIS_boundaries}
      \begin{center}
        \begin{small}
          \begin{sc}
            \resizebox{1\linewidth}{!}{
            \setlength{\tabcolsep}{2pt}
            \begin{tabular}{l c c c c c c c}
                \toprule
                \multirow{2}{*}{Method}                 &   Overall     & \multicolumn{2}{c}{R=0.01}        & \multicolumn{2}{c}{R=0.02} &          \multicolumn{2}{c}{R=0.04}              \\ 
                ~                                       &   mIoU        & mIoU              & F1            & mIoU          & F1                & mIoU              & F1                \\
                \midrule
                CamPoint                                &   83.3        & 74.9              & 84.7          & 79.2          & 89.2              & 81.8              & 92.4              \\
                \midrule
                PointCHR                                & \textbf{86.0} & \textbf{75.3}     & \textbf{86.0} & \textbf{80.4} & \textbf{90.6}     & \textbf{84.1}     & \textbf{93.5}     \\
                \bottomrule
            \end{tabular}}
      \end{sc}
    \end{small}
  \end{center}
  \vskip -0.1in
\end{table}%

\noindent \textbf{Part Segmentation on ShapeNetPart Dataset.}
The ShapeNetPart \cite{yi_scalable_2016} dataset is a widely used benchmark for 3D part segmentation, consisting of 16 object categories with annotated part labels.
We follow the standard evaluation protocol, reporting the mIoU for both class-level and instance-level segmentation. 
As shown in \cref{tab:ShapeNetPart}, the mIoU (\%) for all classes (Cls.) and instances (Ins.) are reported. PointCHR attains SOTA results, achieving an instance mIoU of 87.0\% and a class mIoU of 85.7\%.
This indicates that PointCHR effectively captures fine-grained geometric subtleties, particularly at part transition regions.
These transition zones are geometrically characterized by high curvature and discontinuity, representing the hard samples where Euclidean methods often suffer from feature over-smoothing.
By leveraging the exponential expansion of hyperbolic space, our method preserves the distinct manifold structures of adjacent parts, ensuring precise boundary delineation even in objects with complex topology.

\begin{table}[t]
    \caption{Part segmentation results on ShapeNetPart.}
    \label{tab:ShapeNetPart}
    \begin{center}
    \begin{small}
        \begin{sc}
        \begin{tabular}{l c c c c}
        \toprule
            Method                                          &Ref.           & Ins.         & Cls.     \\
            \midrule
            Mamba3D                         & ACMM 24       & 85.7              & 83.6            \\ 
            PCM                             & AAAI 25       & \underline{86.9}  & \underline{85.0}   \\ 
            Point-MoDE                      & IJCAI 25      & 86.5              & 84.6            \\    
            SI-Mamba                        & CVPR 25       & 85.9              & /               \\  
            PointSD                         & ICCV 25       & 86.1              & 84.5            \\  
            Point-PQAE                      & ICCV 25       & 86.1              & 84.6            \\  
            ZigzagPointMamba                & NeurIPS 25    & 85.8              & 84.2            \\
            PointPETL                       & ICML 25       & 86.2              & 84.7            \\
            \midrule    
            PointCHR                        & ~             & \textbf{87.0}     & \textbf{85.7}   \\ 
            \bottomrule
        \end{tabular}
      \end{sc}
    \end{small}
    \end{center}
    \vskip -0.1in
\end{table}

\noindent \textbf{Shape Classification on ScanObjectNN and ModelNet40 Dataset.}
To comprehensively evaluate the generalization capability of PointCHR, we conduct experiments on two distinct benchmarks: ModelNet40~\cite{wu_3d_2015}, which provides synthetic CAD models for assessing representation capability on canonical shapes, and ScanObjectNN~\cite{uy_revisiting_2019}, a real-world dataset containing noisy and occluded scans to verify robustness against geometric imperfections.
In contrast to segmentation tasks which require dense point-wise discrimination, classification relies on capturing the global structural skeleton of objects.
In this context, high-curvature points serve as the most informative structural anchors for distinguishing similar categories, making our curvature-aware approach inherently suitable for this task.

We initiate our evaluation on the synthetic ModelNet40 dataset, with results summarized in \cref{tab:ModelNet40}.
PointCHR achieves an Overall Accuracy (OA) of 93.7\% and a remarkable Mean Accuracy (mAcc) of 92.0\%.
The superior mAcc demonstrates that our method maintains high discriminative power on tail classes, particularly those characterized by intricate geometric structures or limited training samples.
This result confirms that explicitly modeling the long-tail curvature distribution enhances the discriminative power of global shape descriptors, setting a strong foundation for handling more complex data.

Proceeding to the more challenging real-world domain, we report results on hardest ScanObjectNN variant, \texttt{PB\_T50\_RS}, a dataset characterized by severe background clutter and occlusion.
As shown in \cref{tab:ScanObjectNN}, PointCHR achieves a SOTA OA of 92.7\% and mAcc of 91.7\%.
This significant improvement suggests that the proposed CHR is highly robust to geometric noise. 
By expanding the embedding space for high-curvature features, PointCHR effectively distinguishes intrinsic structural details from background noise, validating its efficacy in real-world scenarios.

\begin{table}[t]
    \caption{Classification OA (\%) and mAcc(\%) on ModelNet40.}
    \label{tab:ModelNet40}
      \begin{center}
        \begin{small}
          \begin{sc}
            \begin{tabular}{l c c c c c}
            \toprule
                Method                      & Ref.          &    OA             &mAcc    \\
                \midrule
                LinNet                      & NeurIPS 24    & 93.6             & 91.0               \\ 
                PCM                         & AAAI 25       & 93.4             & 90.7               \\ 
                CamPoint                    & CVPR 25       & 93.6             & \underline{91.3}   \\
                Point-MoDE                  & IJCAI 25      & \underline{93.6} & /                  \\    
                SI-Mamba                    & CVPR 25       & 92.7             & /                  \\  
                PointSD                     & ICCV 25       & 93.7             & /                  \\  
                Point-PQAE                  & ICCV 25       & 92.8             & /                  \\  
                ZigzagPointMamba            & NeurIPS 25    & 93.2             & /                  \\
                \midrule
                PointCHR                    & ~             & \textbf{93.7}    & \textbf{92.0}      \\ 
                \bottomrule
            \end{tabular}
      \end{sc}
    \end{small}
  \end{center}
  \vskip -0.1in
\end{table}%

\begin{table}[t]
    \caption{Classification OA (\%) and mAcc(\%) on ScanObjectNN. The ScanObjectNN results are reported on the PB\_T50\_RS variant.}
    \label{tab:ScanObjectNN}
      \begin{center}
        \begin{small}
          \begin{sc}
            \begin{tabular}{l c c c c c}
            \toprule
                Method                          & Ref.          &    OA             &mAcc    \\
                \midrule
                LinNet                          & NeurIPS 24    & 88.2              & 86.6            \\ 
                KPConvX-L                       & CVPR 24       & 89.3              & 88.1            \\
                Mamba3D                         & ACMM 24       & 91.8              & /               \\ 
                PCM                             & AAAI 25       & 88.1              & 86.6            \\ 
                CamPoint                        & CVPR 25       & \underline{92.1}  & \underline{91.1}\\
                SI-Mamba                        & CVPR 25       & 87.3              & /               \\  
                PointSD                         & ICCV 25       & 90.1              & /               \\  
                ZigzagPointMamba                & NeurIPS 25    & 88.7              & /               \\
                PointPETL                      & ICML 25    & 90.1              & /               \\
                \midrule
                PointCHR                        & ~             & \textbf{92.7}     & \textbf{91.7}   \\ 
                \bottomrule
            \end{tabular}
      \end{sc}
    \end{small}
  \end{center}
  \vskip -0.1in
\end{table}%

\subsection{Ablation Study}

We conduct a component-wise ablation study to evaluate the individual and joint contributions of the core in PointCHR: HST and PCP. Results are reported on ScanObjectNN and S3DIS Area5, as summarized in \cref{tab:Ablation_Component}.

Starting from the baseline model, we observe that incorporating HST yields a consistent performance improvement. This suggests that modeling semantic interactions in hyperbolic space helps enhance feature discriminability, especially for structurally complex regions. Similarly, enabling PCP alone leads to a more pronounced gain on both benchmarks highlighting the importance of curvature-guided capacity reallocation for capturing fine-grained geometric details.

Notably, combining HST and PCP achieves the best overall performance across both tasks. The complementary gains indicate that HST and PCP address different aspects of the representation challenge: while HST improves semantic mixing under hyperbolic geometry, PCP explicitly redistributes representational capacity according to local geometric complexity. Their synergy effectively alleviates representation crowding and leads to more robust and discriminative point cloud representations.

\begin{table}[t]
    \caption{The efficacy of each component in PointCHR.}
    \label{tab:Ablation_Component}
      \begin{center}
        \begin{small}
          \begin{sc}
            \setlength{\tabcolsep}{2.5pt}
              \begin{tabular}{ c c c c c}
                  \toprule
                  \multirow{2}{*}{Baseline}   &\multicolumn{2}{c}{CHR}                            & ScanObjectNN  & S3DIS        \\
                  ~                           & PCP                       & HST                   & OA(\%)        & mIoU(\%)      \\
                  \midrule
                  \checkmark                  &                           &                       & 92.1          & 83.3          \\
                  \checkmark                  &                           &\checkmark             & 92.2          & 84.1          \\
                  \checkmark                  &\checkmark                 &                       & 92.5          & 84.8          \\
                  \checkmark                  &\checkmark                 &\checkmark             & \textbf{92.8} & \textbf{86.0}\\

                  \bottomrule
              \end{tabular}
          \end{sc}
        \end{small}
      \end{center}
  \vskip -0.1in
\end{table}%

\subsection{Generalization Analysis}

To verify the generalization capability and plug-and-play nature of our proposed method, we integrate the CHR into three representative backbones: PointMLP, DeLA, and PointNext-s. As summarized in \cref{tab:generalization_analysis}, the integration of CHR yields consistent performance gains across diverse architectures and tasks. Specifically, on the ScanObjectNN classification benchmark, all backbones exhibit steady improvements in OA. Notably, on the more challenging S3DIS semantic segmentation task, the benefits are even more pronounced, with DeLA and PointNext-s achieving significant mIoU increases of 2.1\% and 1.9\%, respectively. These results demonstrate that CHR is model-agnostic and effectively enhances the feature discriminability of existing methods without requiring architectural modifications.

\begin{table}[t]
    \caption{Generalization analysis of the proposed CHR.}
    \label{tab:generalization_analysis}
      \begin{center}
        \begin{small}
          \begin{sc}
            \setlength{\tabcolsep}{2.5pt}
              \begin{tabular}{ l c c }
                  \toprule
                  \multirow{2}{*}{Method}     & ScanObjectNN              & S3DIS                 \\
                  ~                           & OA(\%)                    & mIoU(\%)              \\
                  \midrule
                  PointMLP                    & 85.4                      & /                     \\
                  PointMLP+CHR                & 86.2(\textbf{+0.8})       & /                     \\ 
                  DeLA                        & 90.4                      & 73.2                  \\
                  DeLA+CHR                    & 90.9(\textbf{+0.5})       & 75.3(\textbf{+2.1})   \\
                  PointNext-s                 & 87.7                      & 63.4                  \\
                  PointNext-s+CHR             & 88.3(\textbf{+0.6})       & 65.3(\textbf{+1.9})   \\
                  \bottomrule
              \end{tabular}
          \end{sc}
        \end{small}
      \end{center}
  \vskip -0.1in
\end{table}%

\subsection{Model Complexity}
We evaluate the trade-off between segmentation accuracy and resource consumption in \cref{tab:complexity_latency}. 
PointCHR outperforms previous vanilla methods, achieving a dominant mIoU of 86.0\%. 
Critically, this top-tier performance is achieved with a lean architecture of only 21.0M parameters. 
Compared with PointNeXt-XL and PTV3, which rely on scaling up parameters to enhance capacity, PointCHR proves that utilizing curvature-aware hyperbolic embeddings is a more parameter-efficient strategy for capturing fine-grained semantics.
Consequently, the proposed model surpasses the previous SOTA, CamPoint, by 2.7\% in mIoU while adding FLOPs from 3.7G to 4.0G.  
This result highlights the efficacy of our design: by rectifying feature crowding in high-curvature regions, we achieve maximizing representational capacity within a constrained parameter budget.

\begin{table}[t]
    \caption{Comparison of segmentation performance, parameter size, computational cost and inference latency on S3DIS Area 5.}
    \label{tab:complexity_latency}
      \begin{center}
        \begin{small}
          \begin{sc}
            \setlength{\tabcolsep}{2.5pt}
              \begin{tabular}{ l c c c c c}
                  \toprule
                  \multirow{2}{*}{Method}     & mIoU                      & Params.               & FLOPs           &Latency\\
                  ~                           & (\%)                      & (M)                   & (G)             & (ms)\\
                  \midrule      
                  PointNeXt-XL                & 70.8                      & 46.1                  & 84.8            & /\\
                  PointVector-XL              & 72.3                      & 24.1                  & 58.5            & /\\
                  PDNet-XXL                   & 72.3                      & 35.6                  & 12.0            & /\\ 
                  PTV3                        & 73.4                      & 46.2                  & 2.7             & 94\\  
                  DeepLA-120                  & 75.7                      & 30.3                  & 24.7            & /\\
                  CamPoint                    & 83.3                      & 15.8                  & 3.7             & 22\\
                  \midrule      
                  PointCHR                    & 86.0                      & 21.0                  & 4.0             & 25\\
                  \bottomrule
              \end{tabular}
          \end{sc}
        \end{small}
      \end{center}
  \vskip -0.1in
\end{table}%

\begin{table*}[t]
\centering
\caption{
Curvature-stratified quantitative results on S3DIS.
PointCHR consistently improves segmentation performance across all curvature regimes, with the most substantial gain observed in the highest-curvature bin. B and nB denote boundary and non-boundary regions.
}
\label{tab:curvature_stratified_quantitative}
\begin{tabular}{lccccc}
\toprule
Curvature Bin & Point Ratio (\%) & $s$ (B / nB) & Baseline mIoU (\%) & PointCHR mIoU (\%) & $\Delta$ \\
\midrule
$[0.00, 0.05)$ & 83.64 & 1.22 / 1.13 & 84.39 & 86.59 & +2.20 \\
$[0.05, 0.10)$ & 10.34 & 1.84 / 1.43 & 76.75 & 80.91 & +4.16 \\
$[0.10, 0.15)$ & 3.89  & 2.13 / 1.61 & 73.05 & 76.78 & +3.73 \\
$[0.15, 0.20]$ & 2.13  & 2.13 / 1.76 & 66.22 & 76.62 & +10.40 \\
\bottomrule
\end{tabular}
\end{table*}

\begin{table*}[t]
\centering
\caption{
Distribution of the learned scaling factor $s$ across curvature bins and spatial regions.
The threshold $s=1.13$ is used as a data-driven reference value, corresponding to the mean scaling factor of low-curvature non-boundary points.
}
\label{tab:scaling_factor_boundary_distribution}
\begin{tabular}{lccccc}
\toprule
Curvature Bin & Point Ratio (\%) & Avg. $s$ (nB) & $(s>1.13)$ nB Ratio (\%) & Avg. $s$ (B) & $(s>1.13)$ B Ratio (\%) \\
\midrule  
$[0.00, 0.05)$ & 83.64          & 1.13 & 19.81 & 1.22 & 24.18 \\
$[0.05, 0.10)$ & 10.34          & 1.43 & 40.64 & 1.84 & 53.12 \\
$[0.10, 0.15)$ & 3.89           & 1.61 & 49.47 & 2.13 & 61.99 \\
$[0.15, 0.20]$ & 2.13           & 1.76 & 55.50 & 2.13 & 63.00 \\
\bottomrule
\end{tabular}
\end{table*}

\subsection{Curvature-stratified Quantitative Analysis}

To examine PointCHR under different geometric regimes, we conduct a curvature-stratified evaluation on S3DIS. Points are partitioned into four curvature intervals, and segmentation performance is reported for each bin. We also report the point ratio of each bin and the learned scaling factors for boundary and non-boundary regions.

As shown in Table~\ref{tab:curvature_stratified_quantitative}, PointCHR consistently improves performance across all curvature bins, with gains increasing from +2.20\% in the lowest-curvature bin to +10.40\% in the highest-curvature bin. Although the highest-curvature points account for only 2.13\% of all points, PointCHR improves their mIoU from 66.22\% to 76.62\%, demonstrating its effectiveness in mitigating Euclidean representation degradation in geometrically complex regions.

\subsection{Scaling Factor and Boundary Distribution}

We also analyze the learned scaling factor $s$ across curvature bins and spatial regions. Since $s$ controls the strength of feature rectification, its distribution reflects whether PointCHR learns an adaptive, geometry-aware transformation instead of applying uniform scaling. This trend is also consistent with the qualitative observations in \cref{fig:mainfold_tail_analysis} (c) and \cref{fig:main_results_vis}. \cref{tab:scaling_factor_boundary_distribution} shows a clear positive correlation between local curvature and the learned scaling factor. As curvature increases, both the average $s$ and the proportion of high-scaling points consistently increase. 

These results demonstrate that PointCHR assigns stronger rectification to geometrically complex regions. Moreover, boundary points consistently receive larger scaling factors than non-boundary points within the same curvature range, indicating that the learned transformation is both curvature-aware and structure-aware. This provides mechanistic evidence that PointCHR improves performance by selectively enhancing high-curvature and structurally critical regions, rather than relying on uniform feature expansion.

\subsection{Quantitative Analysis of Representation Crowding}
To verify whether PointCHR alleviates Representation Crowding, we evaluate two complementary metrics on S3DIS: Effective Dimensionality (ED) for intrinsic representational capacity and classification margin (M) for class separability. As shown in Table~\ref{tab:representation_crowding}, PointCHR consistently improves the classification margin across all curvature bins, indicating clearer class boundaries. In the high-curvature bins $[0.10,0.20]$, PointCHR also increases ED, suggesting enhanced representational capacity in geometrically complex regions. These results confirm that PointCHR effectively alleviates representation crowding by improving both feature capacity and class separability.

\begin{table}[t]
\centering
\caption{
Quantitative verification of representation crowding on S3DIS. Higher values indicate less representation crowding. B and P denote Base and PointCHR, respectively.
}
\label{tab:representation_crowding}
\begin{tabular}{lcccc}
\toprule
Curvature Bin & ED-B & ED-P & M-B & M-P \\
\midrule
$[0.00,0.05)$ & 7.31  & 6.38  & -2.62 & \textbf{-0.21} \\
$[0.05,0.10)$ & 10.23 & 9.86  & -6.12 & \textbf{-0.97} \\
$[0.10,0.15)$ & 9.06  & \textbf{10.05} & -5.14 & \textbf{-0.89} \\
$[0.15,0.20]$ & 7.49  & \textbf{8.99}  & -3.15 & \textbf{-0.64} \\
\bottomrule
\end{tabular}
\end{table}

\section{Conclusion}
This study introduces PointCHR, a novel curvature-aware hyperbolic rectification designed to enhance point cloud representation learning.
By exploiting the exponential expansion properties of hyperbolic manifolds to overcome the capacity limitations of Euclidean space, PointCHR effectively addresses the representation crowding dilemma, exhibiting exceptional discriminative capability in high-curvature geometric details.
Extensive experiments across multiple benchmarks validate the superiority of PointCHR, achieving SOTA performance in semantic segmentation, part segmentation, and shape classification tasks. 





\section*{Acknowledgements}
This work was supported by the Science and Technology Development Fund of Macau Project 0096/2023/RIA2, 0123/2022/A3, 0044/2024/AGJ, 0140/2024/AGJ, the Chinese National Natural Science Foundation Projects 62406320, and the AI Supercomputing Center of Macau University of Science and Technology (MUST). We also acknowledge the technical support provided by the Information Technology Development Office of MUST, as well as the reviewers AVjy, e3C6, v2GW, and 1Vun for their comments and suggestions on the manuscript.



\section*{Impact Statement}
This paper presents work whose goal is to advance the field of Machine Learning. There are many potential societal consequences of our work, none which we feel must be specifically highlighted here.


\bibliography{main}
\bibliographystyle{icml2026}

\end{document}